\documentclass[runningheads]{llncs}
\usepackage{graphicx}
\usepackage{pifont}
\usepackage{epsfig}
\usepackage{graphicx}
\usepackage{amsmath}
\usepackage{amssymb}
\usepackage{enumitem}
\usepackage[pagebackref=true,breaklinks=true,letterpaper=true,colorlinks,bookmarks=false]{hyperref}
\newcommand{\cmark}{\ding{51}}%
\newcommand{\xmark}{\ding{56}}%
\newcommand{\myParagraph}[1]{\noindent \textbf{#1} ---}

\newcommand{\myObjectiveReward}[2]{\textbf{Objective:} #1 \textbf{Reward:} #2}
\newcommand{\myyes}{\cmark}
\newcommand{\myno}{\xmark}
\begin{document}
\title{Deep Reinforcement Learning on a Budget: 3D Control and Reasoning Without a Supercomputer
}
\titlerunning{3D Control and Reasoning Without a Supercomputer}

\author{Edward Beeching \and
Christian Wolf \and
Jilles Dibangoye\and
Olivier Simonin}
\authorrunning{E. Beeching et al.}
%
\institute{INRIA Chroma team, CITI Lab. INSA Lyon, France.
\url{https://team.inria.fr/chroma/en/} 
\email{\{firstname.lastname\}@inria.fr}}
\maketitle              

\begin{abstract}
An important goal of research in Deep Reinforcement Learning in mobile robotics is to train agents capable of solving complex tasks, which require a high level of scene understanding and reasoning from an egocentric perspective.  When trained from simulations, optimal environments should satisfy a currently unobtainable combination of high-fidelity photographic observations, massive amounts of different environment configurations and fast simulation speeds.
In this paper we argue that research on training agents capable of complex reasoning can be simplified by decoupling from the requirement of high fidelity photographic observations. We present a suite of tasks requiring complex reasoning and exploration in continuous, partially observable 3D environments. 
The objective is to provide challenging scenarios and a robust baseline agent architecture that can be trained on mid-range consumer hardware in under 24h. Our scenarios combine two key advantages: (i) they are based on a simple but highly efficient 3D environment (ViZDoom) which allows high speed simulation (12000fps);
(ii) the scenarios provide the user with a range of difficulty settings, in order to identify the limitations of current state of the art algorithms and network architectures.
We aim to increase accessibility to the field of Deep-RL by providing baselines for challenging scenarios where new ideas can be iterated on quickly. We argue that the community should be able to address challenging problems in reasoning of mobile agents without the need for a large compute infrastructure.
Code for the generation of scenarios and training of baselines is available online at the following repository
 \footnote{\href{https://github.com/edbeeching/3d_control_deep_rl}{ https://github.com/edbeeching/3d\_control\_deep\_rl}\label{note1}}.

\end{abstract}

\begin{figure}[t]
\begin{center}
   \includegraphics[width=0.99\linewidth]{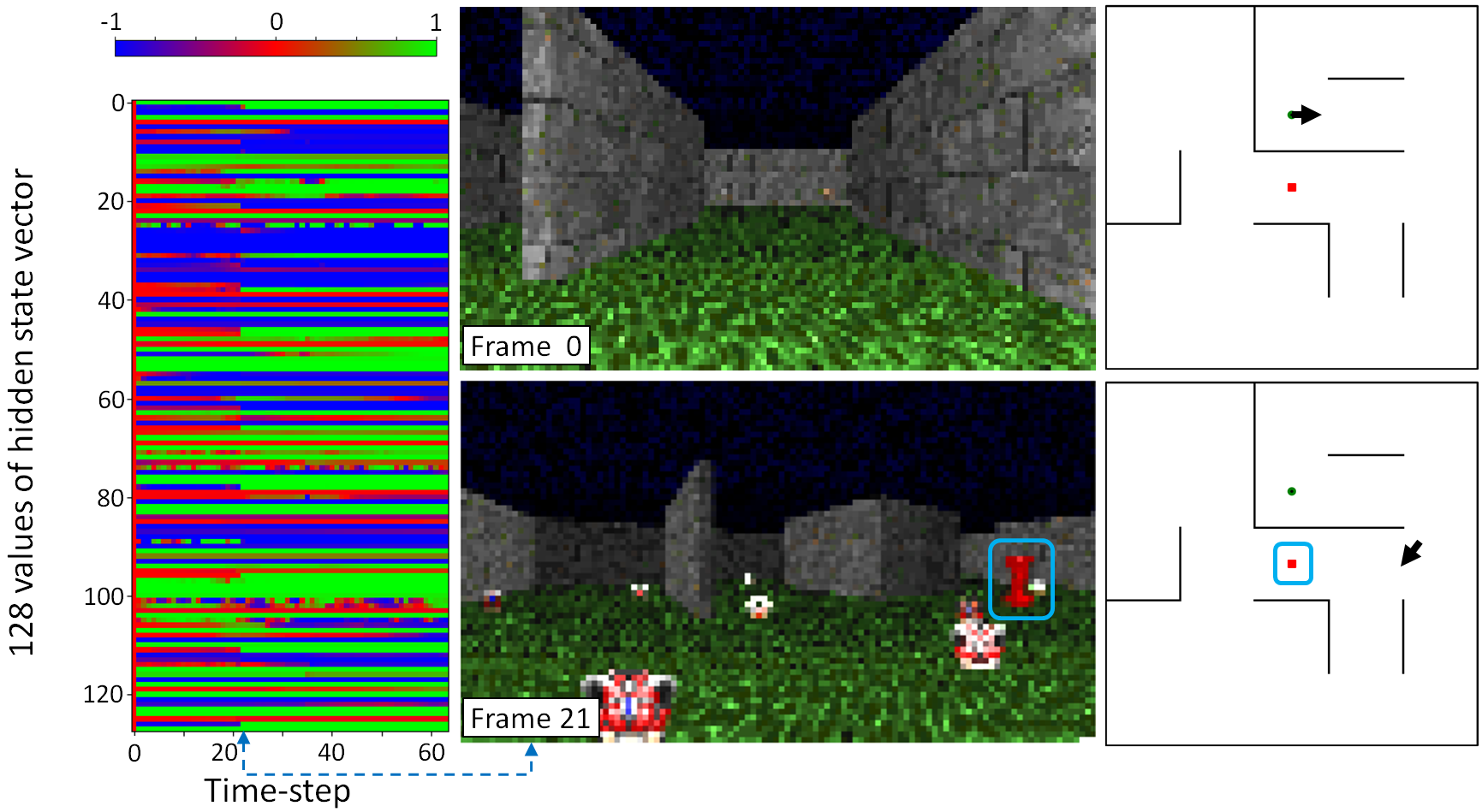}
\end{center}
   \caption{One of the proposed scenarios, called ``\emph{Two color correlation}'': an agent collects items that are the same color as a fixed object at the center of the scenario, which can be either red or green. Left: analysis of the hidden state of a trained agent during an episode. 
    We observe a large change in the activations of the hidden state at step 21, when the fixed object is first observed. Top-center: the agent's viewpoint at step 0. Bottom-center: the agent's viewpoint at step 21, where the agent first observes the fixed object that indicates the objective of the current task (circled in blue). Right: the positions of the agent in the environment.
    }
\label{fig:abstract_figure}
\end{figure}
\section{Introduction}
\noindent
Much of the state of the art research in the field of Deep-RL focuses on problems in fully observable state spaces such as Atari games or in continuous control tasks in robotics, with typical work in grasping \cite{Levine2016End-to-EndPolicies}, locomotion \cite{LillicrapCONTINUOUSLEARNING}, and in joint navigation and visual recognition \cite{Mirowski2016LearningEnvironments,EmbodiedQA}. We address the latter category of problems, which have long term goals and applications in mobile robotics, service robotics, autonomous vehicles and others. We argue that these problems are particularly challenging, as they require the agent to solve difficult perception and scene understanding problems in complex and cluttered environments. Generalization to unseen environments is particularly important, as an agent should be deployable to a new environment, for instance an apartment, house or factory, without being retrained, or with minimal adaptation.
In these scenarios, partial observations of the environment are received from a first person viewpoint. 3D environments require solving the problems of vision, reasoning and long term planning, and are often less reactive than the Atari environments. 

The current state of the art Deep RL algorithms are data hungry, as complex scenarios that necessitate high-level reasoning require hundreds of millions of interactions with the environment in order to converge to a reasonable policy. Even with more sample efficient algorithms such as PPO \cite{Schulman2017ProximalAlgorithms}, TRPO \cite{Wu2017ScalableApproximation} or DDPG \cite{LillicrapCONTINUOUSLEARNING} the sample efficiency is still in the tens of millions. Therefore, even though the question of whether to learn from real interactions or simulated ones is still open, there seems to be a consensus that learning from simulations is currently required at least in parts in order to have access to large scale and high-speed data necessary to train high-capacity neural models. In this respect, an optimal simulated environment should have the following properties:
(i) relatively \textbf{complex reasoning} should be required to solve the tasks;
(ii) the simulated observations are of \textbf{high-fidelity / photo-realistic}, which makes it easier to transfer policies from simulation to real environments;
(iii) \textbf{large-scale}, in a sense that the amount of settings (apartments or houses for home robotics) can be varied. The learned policy should generalize to unseen environments and not encode any particular spatial configuration;
(iv) \textbf{high simulation speed} is required in order to train agents on a massive amount of interactions in a relatively short time.

Unfortunately most of these goals are contradictory. In the past two years many realistic environments have been proposed that offer proxy robotic control tasks. Many of these simulators allow photo-realistic quality rendering but unfortunately only provide a low amount of configurations (apartments) and are slow to simulate even at the lowest of resolutions \cite{Brodeur2017HoME:Environment}, \cite{SavvaMINOS:Environments}, \cite{ai2thor}. This has led to a push to run simulation in parallel on hundreds of CPU cores. 

In this work, we argue that an important step in current research is training agents which are capable of solving tasks requiring fairly complex reasoning, navigation in 3D environments from egocentric observations, interactions with a large amount of objects and affordances etc. Transfer of learned policies to the real world, while certainly being important, can arguably explored in parallel for a large part of our work (as a community). We therefore propose set of highly configurable tasks which can be procedurally generated, while at the same time providing highly efficient simulation speed (see Fig. \ref{fig:abstract_figure} for an example). The work presented aims to be a starting point for research groups who 
wish to apply Deep RL to challenging 3D control tasks that require understanding of vision, reasoning and long term planning, without investing in large compute resources. 

We also provide a widely used training algorithm and benchmarks that can be trained on mid-range hardware, such as 4 CPU cores and a single mid-range GPU (K80, Titan X, GTX 1080), within 24 hours. We provide benchmark results for each scenario in a number of different difficulty settings and identify where there are areas for improvement for future work. We evaluate the generalization performance of the agents on unseen test scenario configurations. 

To summarize, we present the following contributions:
\setlist{leftmargin=*}
\begin{itemize}
\itemsep-0.1em
    \item A benchmark suite of proxy tasks for robotics, which require cognitive reasoning while at the same time being computationally efficient. The scenarios have been built on top of the ViZDoom simulator \cite{Kempka2017ViZDoom:Learning} and run at 12,000 environment interactions per second on limited compute, or 9,600 per second including the optimization of the agent's policy network.
    \item A baseline agent, which solves the standard ViZDoom tasks and is benchmarked against our more challenging and diverse scenarios of varying difficulty settings.
    \item Experiments that demonstrate generalization performance of an agent in unseen scenario configurations.
    \item An analysis of the activation's in the hidden state of a trained agent during an episode.
\end{itemize}
This paper is structured as follows: section \ref{sec:related_work} discusses the advantages and limitations of simulators available in Deep RL, pre-existing benchmarks and recent analyses of the instability and generalization performance of RL. Section \ref{sec:scenarios} describes each scenario and the reasoning we expect the agent to acquire. In section \ref{sec:baseline_agent} we describe the baseline agent's architecture, key components of the training algorithm and the experimental strategy. Results and analysis are shown in section \ref{sec:results} and conclusions made in section \ref{sec:conclusions}.


\section{Related Work}\label{sec:related_work}
\begin{table}[t]
\caption{\label{tab:simulators} Comparison of the first person 3D environment simulators available in the RL community. Frames per second (FPS) is shown for a single CPU core.}
\begin{tabular}{lccccrr}
\hline
\multicolumn{1}{c}{\textbf{Simulator}} & \textbf{Difficulty} & \textbf{Duration} & \textbf{Extendable} & \textbf{Realistic} & \multicolumn{1}{c}{\textbf{Scenarios}} & \multicolumn{1}{c}{\textbf{FPS}} \\ \hline
VizDoom\cite{Kempka2017ViZDoom:Learning} & Simple & Short & \myyes & \myno & 8 & 3000+ \\
Deepmind Lab\cite{Beattie2016DeepMindLab} & Complex & Long & \myyes & \myno & 30 & 200+ \\
HoME\cite{Brodeur2017HoME:Environment} & Simple & Short & \myno & \myyes & 45000 & 200+ \\
Habitat\cite{habitat19arxiv} & Simple + text & Short & \myno & \myyes & 45662 & 200+ \\
Minos\cite{SavvaMINOS:Environments} & Simple & Short & \myno & \myyes & 90 & 100+ \\
Gibson\cite{xiazamirhe2018gibsonenv} & Simple + text & Short & \myno & \myyes & 572 & 50+ \\
Matterport\cite{mattersim} & Simple + text & Short & \myno & \myyes & 90 & 10+ \\
AI2-thor\cite{ai2thor} & Simple & Short & \myno & \myyes & 32 & 10+ \\
\end{tabular}
\end{table}
\myParagraph{Deep Reinforcement Learning}
In recent years the field of Deep Reinforcement Learning (RL) has gained attention with successes on board games  \cite{SilverMasteringSearch} and Atari Games \cite{Mnih2015Human-levelLearning}. One key component was the application of deep neural networks \cite{Lecun1998Gradient-BasedMethod} to frames from the environment or game board states. Recent works that have applied Deep RL for the control of an agent in 3D environments such as maze navigation are \cite{Mirowski2016LearningEnvironments} and \cite{Jaderberg2016ReinforcementTasks} which explored the use of auxiliary tasks such as depth prediction, loop detection and reward prediction to accelerate learning. Meta RL approaches for 3D navigation have been applied by \cite{Wang2016LearningLearn} and \cite{Lample} also accelerated the learning process in 3D environments by prediction of tailored game features. There has also been recent work in the use of street-view scenes in order train an agent to navigate in city environments \cite{Kayalibay2018NavigationMaps}. In order to infer long term dependencies and store pertinent information about the environment; network architectures typically incorporate recurrent memory such as Gated Recurrent Units \cite{ChungGatedNetworks} or Long Short-Term Memory \cite{Hochreiter1997LONGMEMORY}.\newline
The scaling of Deep RL has produced some impressive results, such as in the IMPALA\cite{Espeholt2018IMPALA:Architectures}
architecture which successfully trained an agent that can achieve human level performance in all 30 of the 3D partially observable DeepMind Lab \cite{Beattie2016DeepMindLab} tasks; accelerated methods such as in \cite{Stooke2018AcceleratedLearning} which solve many Atari environments in tens of minutes and the recent achievement in long term planning on the game of Dota by the OpenAI5 \cite{OpenAI_dota} system have shown that Deep RL agents can be trained with long horizons and a variety of settings. All of these systems represent the state of the art in the domain of Deep RL, the downside is that the hardware requirements required to train the agents make this level of research prohibitive for all but the largest of institutions.
\begin{figure}[t] \centering
  \includegraphics[width=0.9\linewidth]{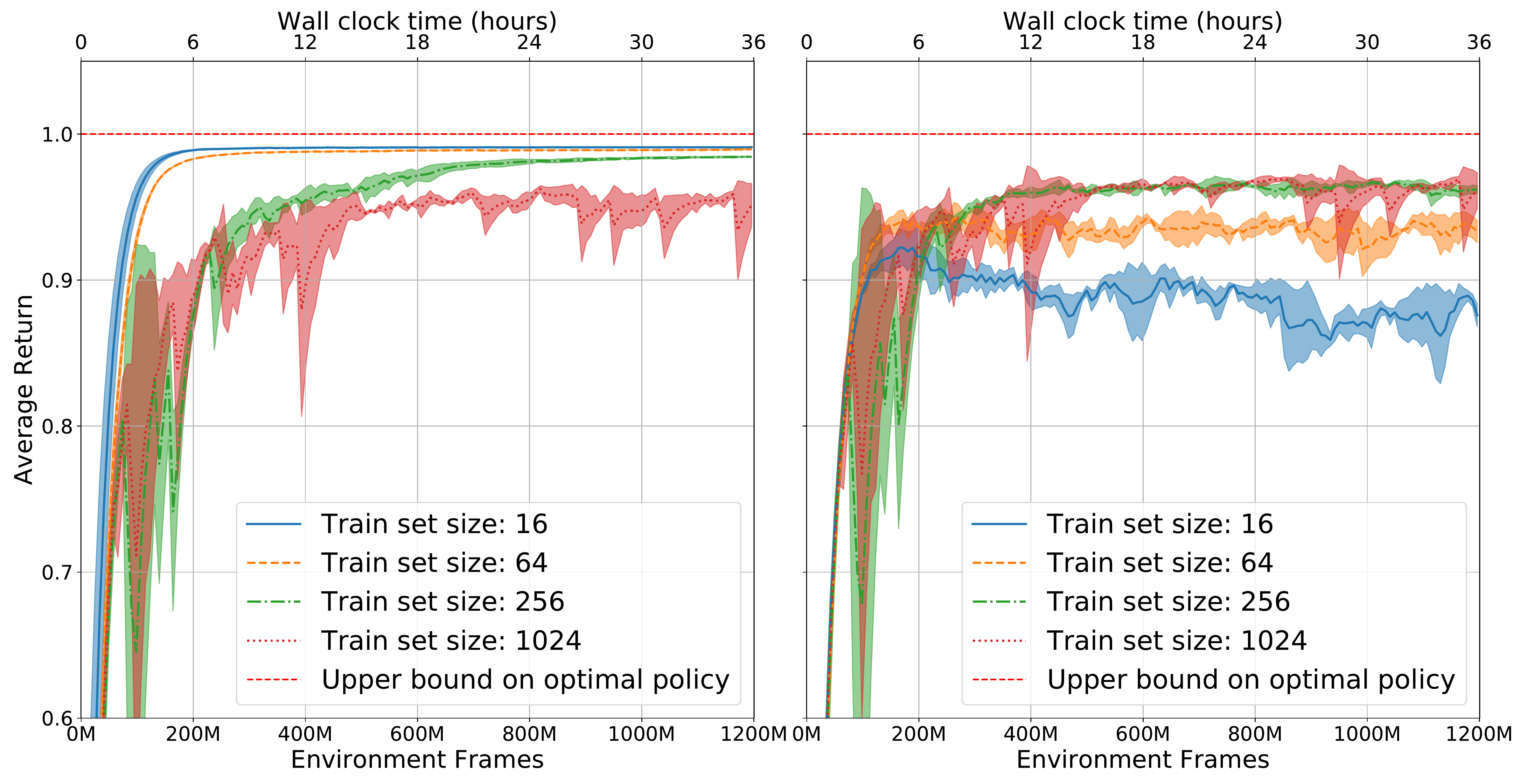}
  \caption{Generalization to unseen maze configurations: the average return and std. of three separate experiments trained on the labyrinth scenario with training set sizes of 16, 64, 256 and 1024 mazes, evaluated on the training set (left) and a held out set of 64 mazes (right). We observe that the training sets of size 16 and 64 quickly overfit to the training data and their generalization performance initially improves and then degrades over time. With 256 and 1024 maze configurations, the agent's test set performance plateaus at the same level.}
\label{fig:gen_tests}
\end{figure}

\myParagraph{Environments}
Beyond the environments available in the Gym framework \cite{Brockman2016OpenAIGym} for Atari and continuous control tasks, there are numerous 3D simulators available with a range advantages and disadvantages. Before starting this work, we investigated the limitations of each simulator, which are summarized in table \ref{tab:simulators}, as we focus on proxies for in home mobile robots we exclude comparisons with car and drone simulators such as CARLA \cite{Dosovitskiy17} and Sim4CV \cite{2017arXiv170805869M}. In choosing the simulator, speed was of utmost importance as we did not want to starve the GPU of data. The fidelity and realism of the simulator was not of great concern, as the intention was to run the simulator at a low resolution where fine detail cannot be captured. We chose ViZDoom as it is orders of magnitude faster than its competitors. The downside of ViZDoom is that there are only 8 scenarios available which are limited in scope and cannot be used to evaluate the generalization performance of an agent. For this reason we have constructed an ever growing collection of scenarios that aim to test navigation, reasoning and memorization. Four of the scenarios have been included in the proposed benchmark. 

\myParagraph{Other benchmarks}
There are a number of open source benchmark results and libraries available in the field of Deep RL. The most popular being OpenAI baselines \cite{baselines} which focuses on the Atari and Mujuco continuous control tasks. There are several libraries which such as Ray \cite{Moritz} and TensorForce \cite{schaarschmidt2017tensorforce} that provide Open Source implementations of many RL algorithms. Unfortunately these implementations are applied in environments such Atari and Mujuco continuous control and standard benchmark results are not available for Deep RL applied in 3D environments. 

To our knowledge, the most similar open-source benchmark of Deep RL applied to 3D scenarios is IMPALA \cite{Espeholt2018IMPALA:Architectures}. Although an impressive feat of engineering, the downside of this solution is the high compute requirement with parallel environments running on up to 500 compute cores. Our scenarios provide comparable levels of reasoning and run on limited hardware due a highly optimized simulator.

\begin{figure}[t]
\begin{center}
  \includegraphics[width=0.99\linewidth]{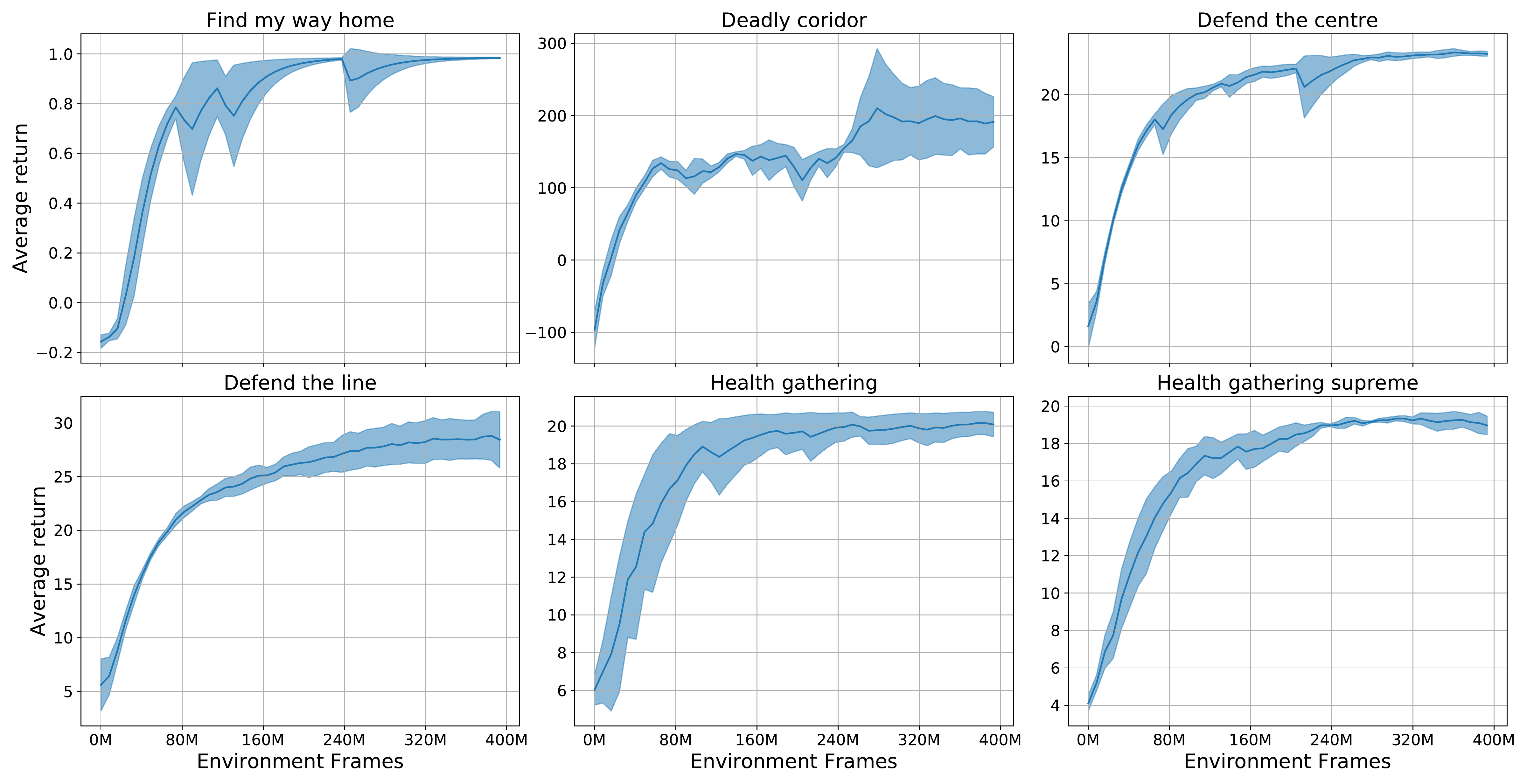}
\end{center}
  \caption{Training curves for the ViZDoom standard scenarios, shown are the mean and standard deviation for three independent experiments conducted for each scenario.}
\label{fig:standard_scenarios}
\end{figure}
\myParagraph{Algorithmic instability and generalization}
In the last year work has been published on the analysis of the generalization and the often under-documented challenge of algorithmic stability in the field of Deep RL. Much of the focus in Deep RL is training an agent to master a specific task, which is in contrast to supervised learning where data is held out in order to evaluate the performance on similar but unseen data drawn for the same underlying distribution. Recent work has studied over-fitting in Deep RL in 2D gridworld scenarios \cite{Zhang2018ALearning} which motivated us to repeat similar experiments in 3D scenarios with separate training and test datasets. An experimental study \cite{Henderson2017DeepMatters} highlights the sensitivity of Deep RL to weight initialization and hyper parameter configurations which are often overlooked when results are discussed. In the benchmark details section, we explicitly state our training strategy and evaluation procedures.

\section{Scenarios and required reasoning}\label{sec:scenarios}
\noindent
We have designed scenarios that aim to: \\
\noindent
\myParagraph{Be proxies for mobile robotics} the tasks represent simplified versions of real world tasks in mobile and service robotics, where an agent needs to act in an initially unknown environment, discovering key areas and objects. 
This also requires the tasks to be partially observable and to provide 3D egocentric observations. \\
\noindent
\myParagraph{Test spatial reasoning} the scenarios require the agent to explore the environment in an efficient manner and to learn spatial-temporal regularities and affordances. The agent needs to autonomously navigate, discover objects, eventually store their positions for later use if they are relevant, their possible interactions, the eventual relationships between the objects and the task at hand. Semantic mapping is a key feature in these scenarios. \\
\noindent
\myParagraph{Discover semantics from interactions} while solutions exist for semantic mapping and semantic SLAM \cite{6094648}, we are interested in tasks where the semantics of objects and their affordances are not supervised, but defined through the task and thus learned from reward. \\
\noindent
\myParagraph{Generalize to unseen environments} related to being proxies for robotics, the trained policies should not encode knowledge about a specific environment configuration\footnote{In service robotics we require a robot to work in an unseen home.}. The spatial reasoning capabilities described above should thus be learned such that an agent can make a difference between affordances, which generalize over configurations, and between spatial properties which need to be discovered in each instance/episode. We thus generate a large number of different scenarios in a variety of configurations and difficult settings and split them into different subsets, evaluating the performance on test data.

\begin{figure}[t]
    \centering
    \begin{minipage}{0.49\textwidth}
        \centering
        \includegraphics[width=1.0\textwidth]{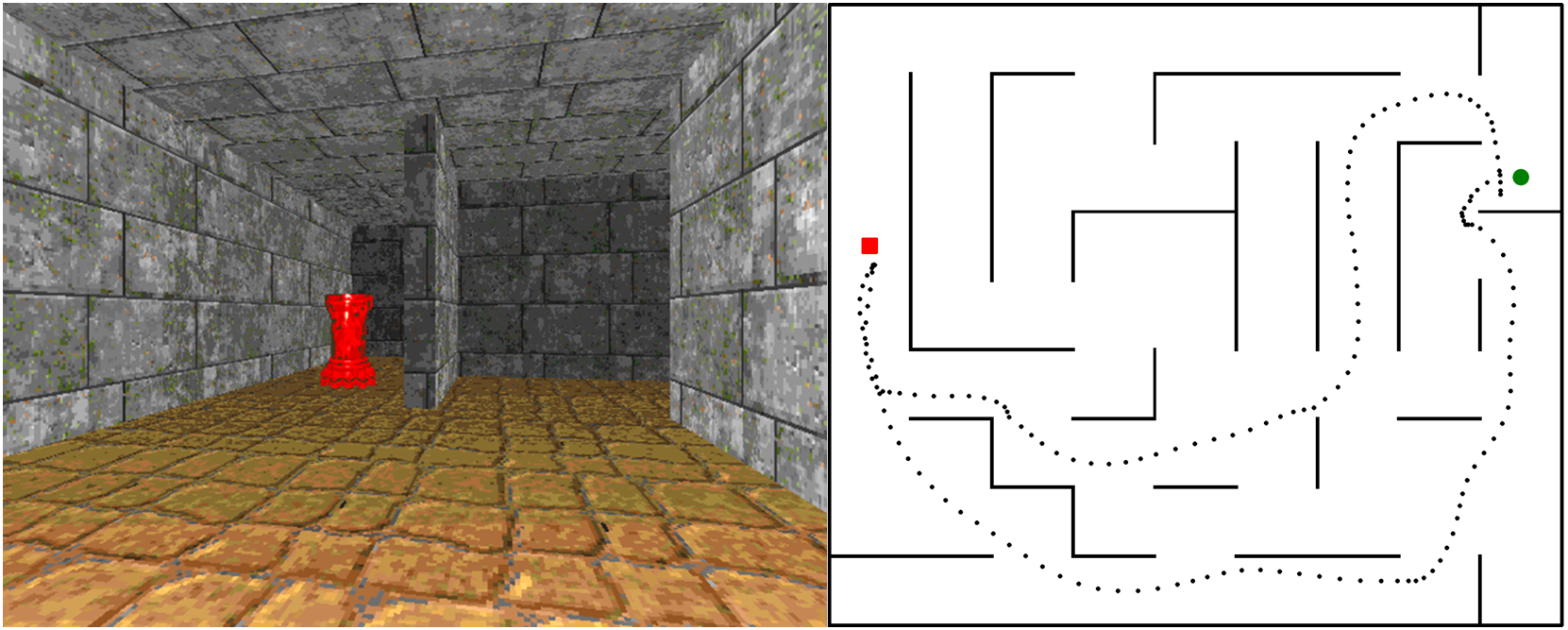} 
        \caption{\label{fig:scenario_find_return}The ''\emph{Find and return}'' scenario. Left: An agent's egocentric viewpoint. Right: A top down view; the agent starts at the green object, finds the red object and then returns to the entry.}
    \end{minipage}\hfill
    \begin{minipage}{0.49\textwidth}
        \centering
        \includegraphics[width=1.0\textwidth]{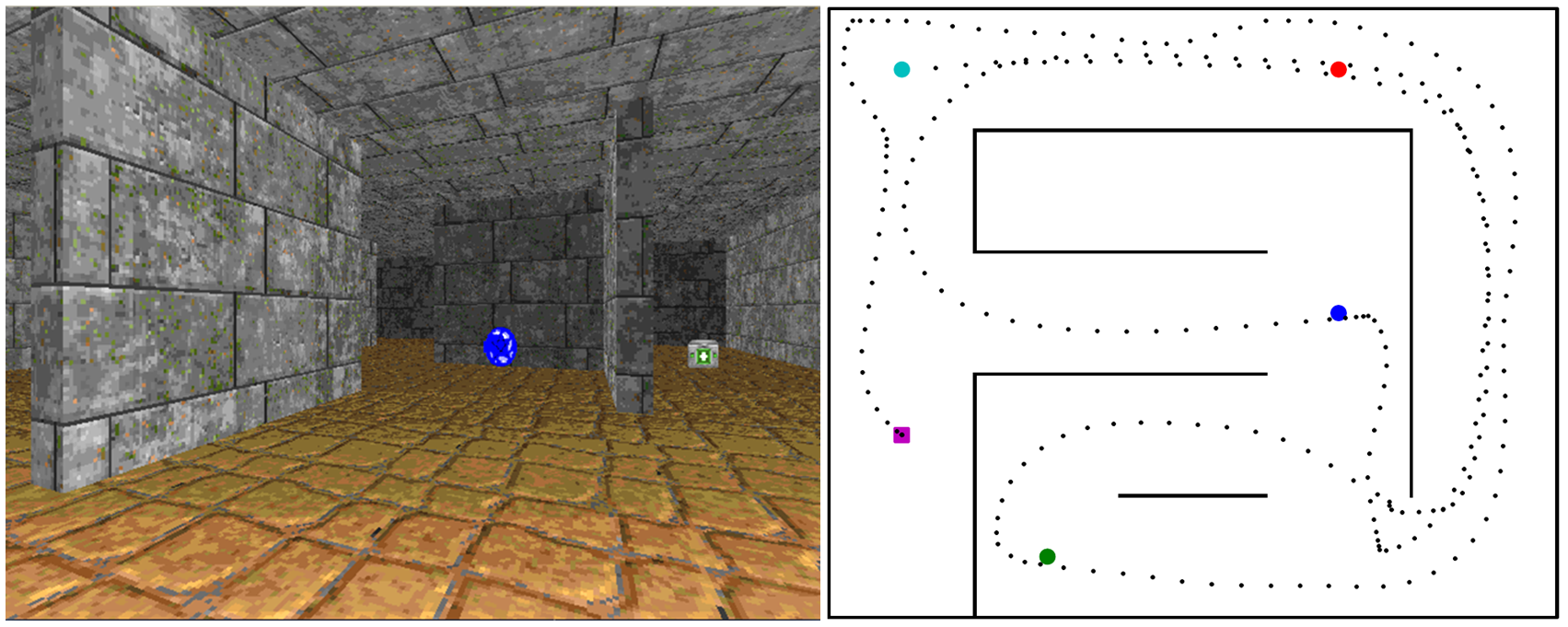} 
        \caption{\label{fig:scenario_kitem}The ordered $K-$item scenario, where $K{=}4$. Left: The agent's view of the scenario. Right: a top-down view of an episode where the agent collects the 4 items in the correct order.}
    \end{minipage}
\end{figure}

We propose a set of four tasks called ``\emph{Labyrinth}'',  ``\emph{Find and return}'',  ``\emph{Ordered K-item}'' and  ``\emph{Two color correlation}''. They are build on top of the ViZDoom simulator\cite{Kempka2017ViZDoom:Learning} and involve an agent moving in a maze with various objects. They have continuous partially observable state spaces with a discrete action space including $\{$\emph{forward, backward, turn left, turn right,}$\dots\}$. The mazes were generated with a depth first search, further interconnections were added by randomly removing 40\% of the walls. For the two color scenario the percentage of walls removed was varied in order to evaluate performance with changing scenario complexity.
For each scenario, 256 configurations were used for training and 64 for testing. This was determined with a study on the labyrinth scenario (described in section \ref{sec:labyrinth}) with the baseline agent described in section \ref{sec:baseline_agent}. We concluded that 256 scenarios are sufficient to train a general policy, shown in Fig. \ref{fig:gen_tests}. 

We chose to run the simulator at a resolution of size 64$\times$112 (h$\times$w), which is comparable in size to the state of the art \cite{Espeholt2018IMPALA:Architectures} of 72$\times$96. We observed that a wider aspect ratio increases the field of view of the agent and enables each observation to provide more information about the environment for a similar pixel budget. We run the same architeture and hyperparameter settings on the standard ViZDoom scenarios, results are shown in Figure \ref{fig:standard_scenarios}.
\begin{figure*}[t] \centering
  \includegraphics[trim={0 8.5cm 0 0 }, clip, width=1.0\linewidth]{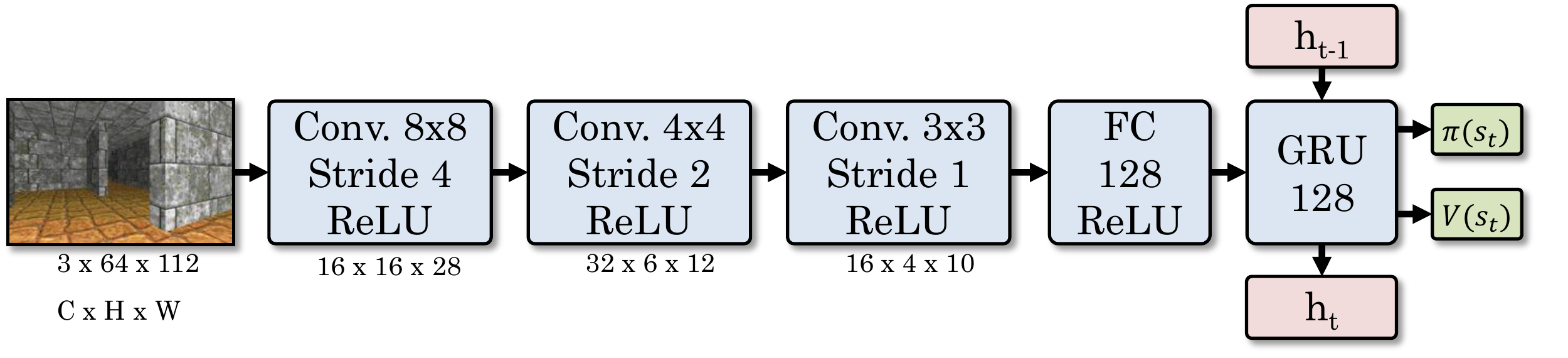}
  \caption{The benchmark agent's architecture}
\label{fig:baseline_arch}
\end{figure*}

\myParagraph{Labyrinth}
\label{sec:labyrinth}
\noindent
The easiest scenario, labyrinth, tests the ability of an agent to efficiently explore an environment and learn to recognize an exit object. A typical maze scenario, with wall configurations created with procedural generation.
Rewards are sparse with long time dependencies between actions and rewards.
\myObjectiveReward
{Find the exit of a randomly generated maze of size n$\times$n in the shortest time.}
{+1 for finding the exit, -0.0001 for each time-step, limited to 2100 steps.}

\myParagraph{Find and return}
\noindent
A general and open source implementation of the ``Minotaur'' scenario detailed in \cite{Parisotto2017NeuralLearning}, it tests the spatial reasoning of an agent by encouraging it to store in it's hidden state the configuration of an environment, so that it can quickly return to the starting point once the intermediate object has been discovered (see Fig. \ref{fig:scenario_find_return}).
\myObjectiveReward
{Find an object in a maze and then return to the starting point.}
{+0.5 for finding the object, +0.5 for return to the starting point, -0.0001 for each time-step, limited to 2100 steps.}

\myParagraph{Ordered k-item}
\noindent
In this more challenging scenario, an agent must find $K$ objects in a fixed order. It tests three aspects of an agent, it's ability to: explore the environment in an efficient manner, learn to collect items in a predefined order and store as part of it's hidden state where items were located so they can be retrieved in the correct order. Shown in Fig. \ref{fig:scenario_kitem}.
\myObjectiveReward
{Collect k items in a predefined order.}
{+0.5 for finding an object in the right order, -0.25 for finding an object in the wrong order, which also ends the scenario, -0.0001 for each time-step, limited to 2100 steps.}
\begin{figure}[t]
\begin{center}
  \includegraphics[width=0.99\linewidth]{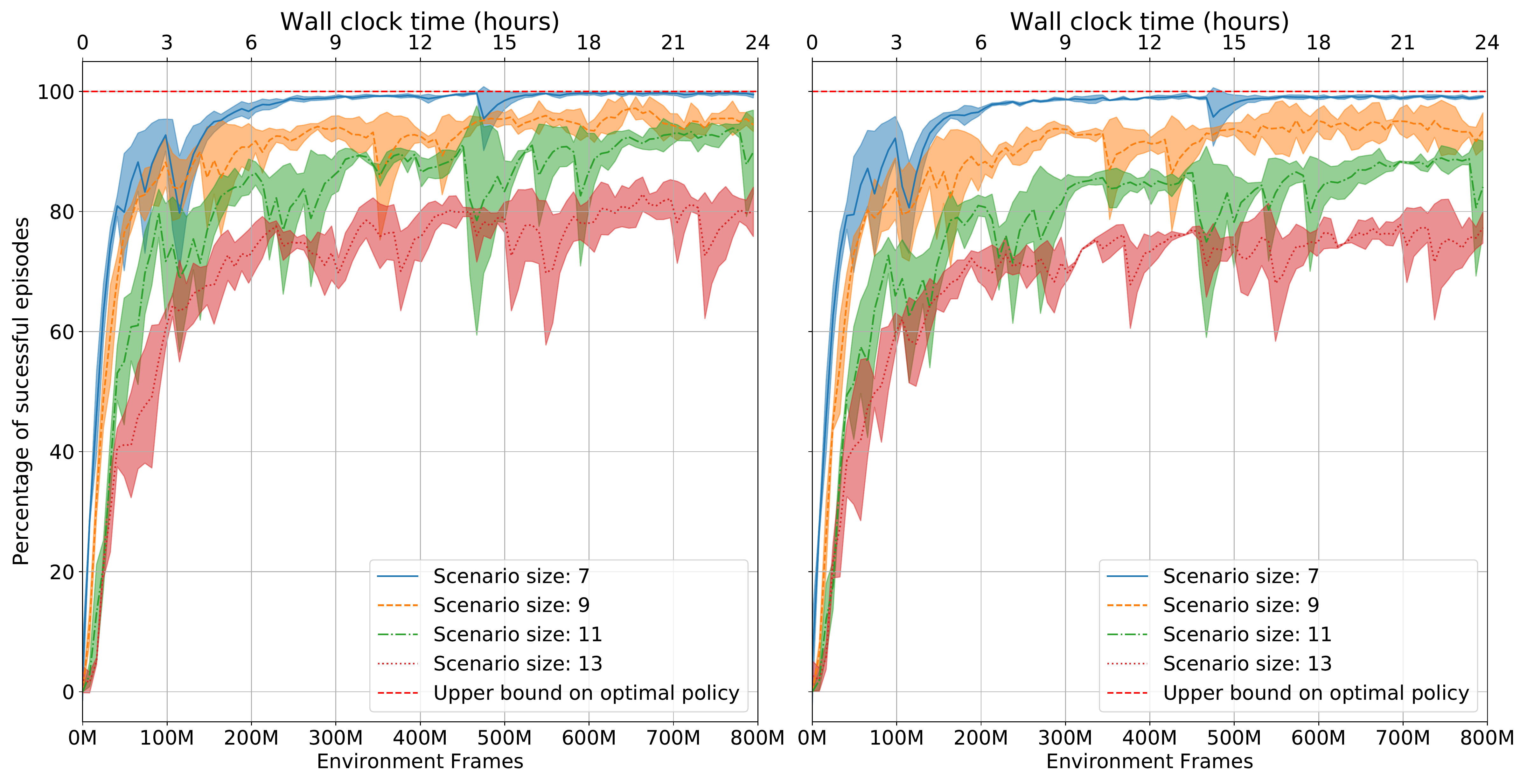}
\end{center}
  \caption{Results from the ''\emph{Labyrinth}'' scenario of sizes $[7,9,11,13]$, three experiments were conducted per size. Shown are the mean and std.\ of the percentage of successful episodes evaluated on the training data (left) and the test data (right). The performance of the benchmark agent diminishes as the size of the scenario increases.}
\label{fig:laby_results}
\end{figure}

\myParagraph{Two color correlation}
\noindent
Potentially the most challenging of the scenarios, here the agent must infer the correlation between a large red or green colored object in the scenario and the correct items to collect. The scenario tests an agent's ability to explore the environment quickly in order to identify the current task and then to store the current task in it's hidden state for the duration of the scenario.
\myObjectiveReward
{Collect the correct color objects in a maze full of red and green objects, the task is to pick up the object which is the same color as a totem at the centre of the environment. The agent also starts with 100 health, collecting the correct item increases health by 25, collecting the wrong item decreases the agent's health by 25, if the agent's health is below 0, the scenario is reset. The agent's health also decays by 1 each time-step to encourage exploration.}
{+0.1 for the correct item, -0.01 per time-step.}

\section{Benchmark agent details}\label{sec:baseline_agent}
\noindent
We apply a popular model-free policy gradient algorithm, Advantage Actor Critic (A2C) \cite{Mnih2016AsynchronousLearning}, implemented in the PyTorch framework by \cite{pytorchrl}. 
A2C optimizes the expected discounted future return $R_t = {\sum_{t=0}^{T}\gamma^t r_t}$ over trajectories of states $s_t$, action $a_t$ and reward $r_t$ triplets collected during rollouts of a policy in the environment. The algorithm achieves this by optimizing a policy $\pi(.|s_t;\theta)$ and a value function $V^\pi(s_t;\theta)$. Both the policy and value function are represented by neural networks with shared parameters $\theta$ that are optimized by gradient descent.
\begin{table}[t]
\centering
\caption{\label{tab:results}Summary of results on the training and test sets for the four scenarios in a variety of difficulty settings.}
\label{my-label}
\begin{tabular}{cccccccl}
\cline{1-7}
\multicolumn{3}{c}{\textbf{Labyrinth}} & \multicolumn{1}{l}{\textbf{}} & \multicolumn{3}{c}{\textbf{Ordered K-item}} &  \\ \cline{1-3} \cline{5-7}
\textbf{Size} & \textbf{Train} & \textbf{Test} & \textbf{} & \textbf{\# items} & \textbf{Train} & \textbf{Test} &  \\
7 & 99.8 $\pm$ 0.2 & 99.5 $\pm$ 0.2 &  & 2 & 0.968 $\pm$ 0.004 & 0.932 $\pm$ 0.012 &  \\
9 & 98.1 $\pm$ 0.8 & 97.4 $\pm$ 1.3 &  & 4 & 0.910 $\pm$ 0.012 & 0.861 $\pm$ 0.015 &  \\
11 & 95.2 $\pm$ 0.7 & 90.1 $\pm$ 1.0 &  & 6 & 0.577 $\pm$ 0.024 & 0.522 $\pm$ 0.023 &  \\
13 & 84.1 $\pm$ 1.8 & 79.8 $\pm$ 1.9 &  & 8 & 0.084 $\pm$ 0.071 & 0.083 $\pm$ 0.070 &  \\ \cline{1-7}
\multicolumn{1}{l}{} & \multicolumn{1}{l}{} & \multicolumn{1}{l}{} & \multicolumn{1}{l}{} & \multicolumn{1}{l}{} & \multicolumn{1}{l}{} & \multicolumn{1}{l}{} &  \\ \cline{1-7}
\multicolumn{3}{c}{\textbf{Find Return}} & \multicolumn{1}{l}{\textbf{}} & \multicolumn{3}{c}{\textbf{Two colors}} &  \\ \cline{1-3} \cline{5-7}
\textbf{Size} & \textbf{Train} & \textbf{Test} & \textbf{} & \textbf{Complexity} & \textbf{Train} & \textbf{Test} &  \\
7 & 98.7 $\pm$ 0.3 & 95.1 $\pm$ 0.2 &  & 1 & 1903 $\pm$ 27 & 1941 $\pm$ 4 &  \\
9 & 87.0 $\pm$ 1.9 & 81.6 $\pm$ 1.8 &  & 3 & 1789 $\pm$ 24 & 1781 $\pm$ 24 &  \\
11 & 70.9 $\pm$ 1.6 & 64.0 $\pm$ 1.6 &  & 5 & 1436 $\pm$ 158 & 1432 $\pm$ 161 &  \\
13 & 57.7 $\pm$ 0.8 & 52.9 $\pm$ 3.7 &  & 7 & 1159 $\pm$ 126 & 1128 $\pm$ 140 &  \\ \cline{1-7}
\end{tabular}
\end{table}

The A2C algorithm is an extension of REINFORCE \cite{REINFORCE} where the variance of the updates are reduced by weighting by the advantage $A(a, s_t){=}R_t{-}V(s_t)$. The objective function that is optimized is detailed in equation \ref{eq:loss_funct}, the $\lambda$ factor weights the trade-off between the policy and value functions, and is typically 0.5 for most A2C implementations. 
\begin{equation}\label{eq:loss_funct}
L(\theta) = -\log \pi(a_t|s_t;\theta)A(a_t, s_t) + \lambda (R_t-V(s_t;\theta))^2 
\end{equation}
An optional entropy term is included to encourage exploration by penalizing peaky, low entropy distributions. We found the weighting of the entropy term to be key during hyper-parameter tuning.

The algorithm was run in a batched mode with 16 parallel agents sharing trajectories of over 128 environment steps, with discounted returns bootstrapped from value estimates for non-terminal states. The forward and backward passes were batched and computed on a GPU for increased efficiency.
The downside of this setup is that the slowest simulator step will bottleneck the performance, as the chosen simulator is exceptionally fast we found in practice that this did not have a large impact. We achieved a rate of training of approximately 9,200 environment frames per second. This includes a 4 step frame skip where physics are simulated but the environment is not rendered, so on average we simulate and process 2,300 observations per second. The training efficiency was benchmarked on Xeon E5-2640v3 CPUs, with 16GB of memory and one NVIDIA GK210 GPU.
The benchmark agent's network architecture is a 3 layer CNN similar to that used in \cite{Mnih2015Human-levelLearning} with 128 GRU to capture long term information, shown in Fig. \ref{fig:baseline_arch}. 
Three independent experiments were undertaken for each scenario and difficulty combination in order to evaluate algorithmic stability. Agents were trained for 200M observations from the environment, which is equal to 800M steps with a frame skip of 4. For the more complex variants of the scenarios we do not train the networks to convergence, as training was capped at 200M observations. 

All experiments were conducted with the same hyper-parameter configuration, which was observed to be stable for all 4 custom scenarios and also the default scenarios provided with the ViZDoom environment.
The hyper-parameters were chosen after a limited sweep across learning rate, gamma factor, entropy weight and n-step length (the number of environment steps between network updates) on the labyrinth scenario. Weights were initialized with orthogonal initialization and the appropriate gain factor was applied for each layer.

\section{Results and Analysis}\label{sec:results}
\begin{figure}[t]
\begin{center}
  \includegraphics[width=0.99\linewidth]{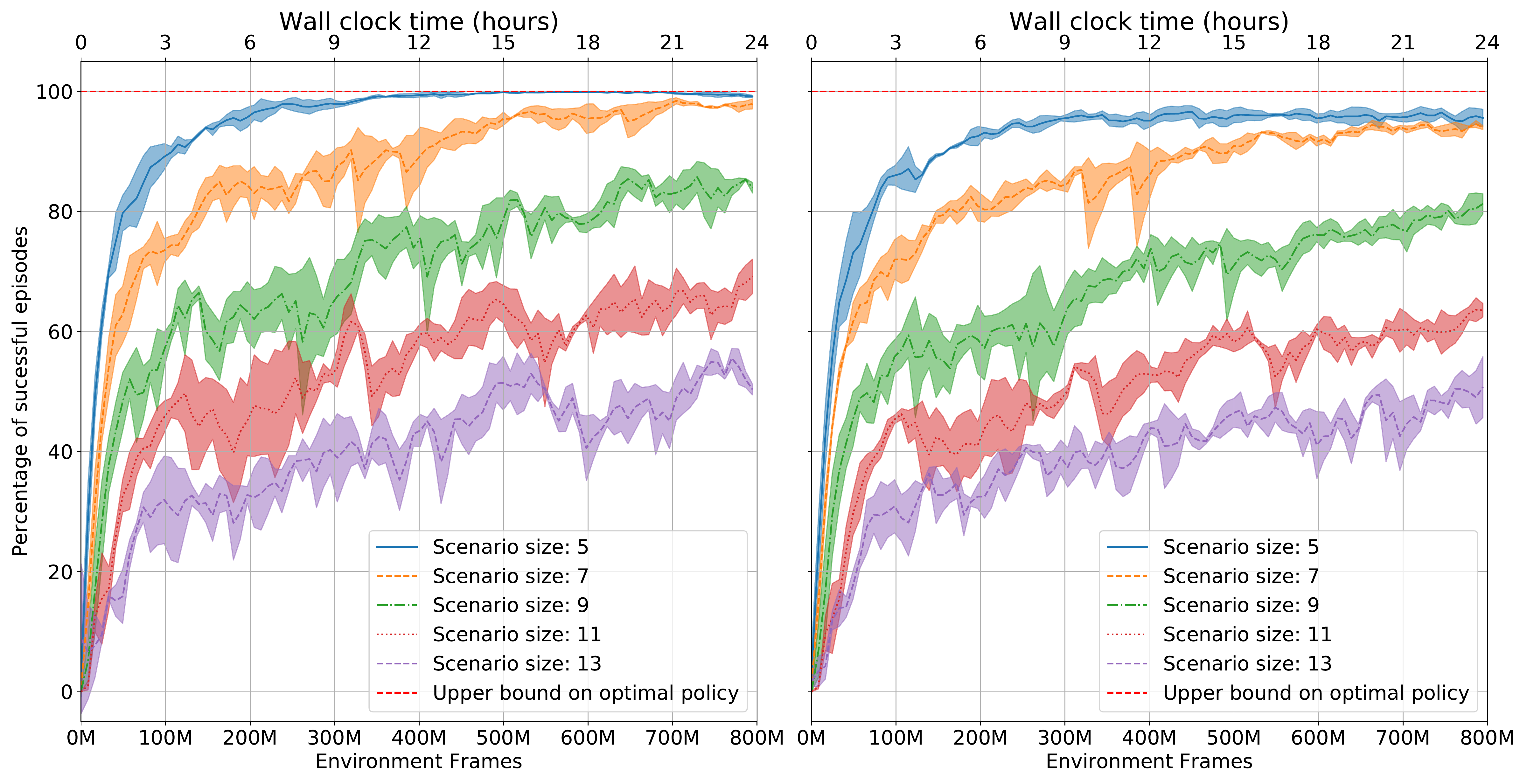}
\end{center}
  \caption{Results from the ''\emph{Find and return}'' scenario of sizes $[5,7,9,11,13]$, three experiments were conducted per size. Displayed are the mean and std.\ of the percentage of successful episodes evaluated on training data (left) and test data (right).}
\label{fig:find_return_results}
\end{figure}
For each scenario there are a variety of possible difficulty settings. We have explored a range of simple to challenging configuration options and evaluated the agent's performance when trained for 200 M observations from the environment. 

For each possible sub-configuration, 3 independent experiments were conducted with different seeds for the weight initialization in order to get an estimate of the stability of the training process. Results of the agent's mean performance on training and test datasets are shown with errors of one standard deviation across the 3 seeds. A summary of all results is included in table \ref{tab:results}. For each plot in the next section, we show a measure of the agent's performance such as average return or percentage of successful episodes on the y-axis and the number of environment frames on the x-axis. 
\subsection{Results}
\noindent
\textbf{Labyrinth:} Experiments were conducted on 4 sizes of the find and return scenario ranging from 7x7 to 13x13, results of the agent's performance shown in Fig. \ref{fig:laby_results}. We observe that the performance of the benchmark plateaus near to the optimal policy for the scenarios of size 7 and 9. Reasonable performance is achieved with the size 9 scenario and performance is reduced for the size 13 scenario, which is challenging for the agent. 

\noindent
\textbf{Find and return:} Experiments were conducted on 5 sizes of the find and return scenario ranging from 5x5 to 13x13, results of the agent's performance are shown in Fig. \ref{fig:find_return_results}.
We observe reasonable agent performance on the smaller scenario configurations and scenarios of size 11+ are currently challenging.

\begin{figure}[t]
\begin{center}
  \includegraphics[width=0.99\linewidth]{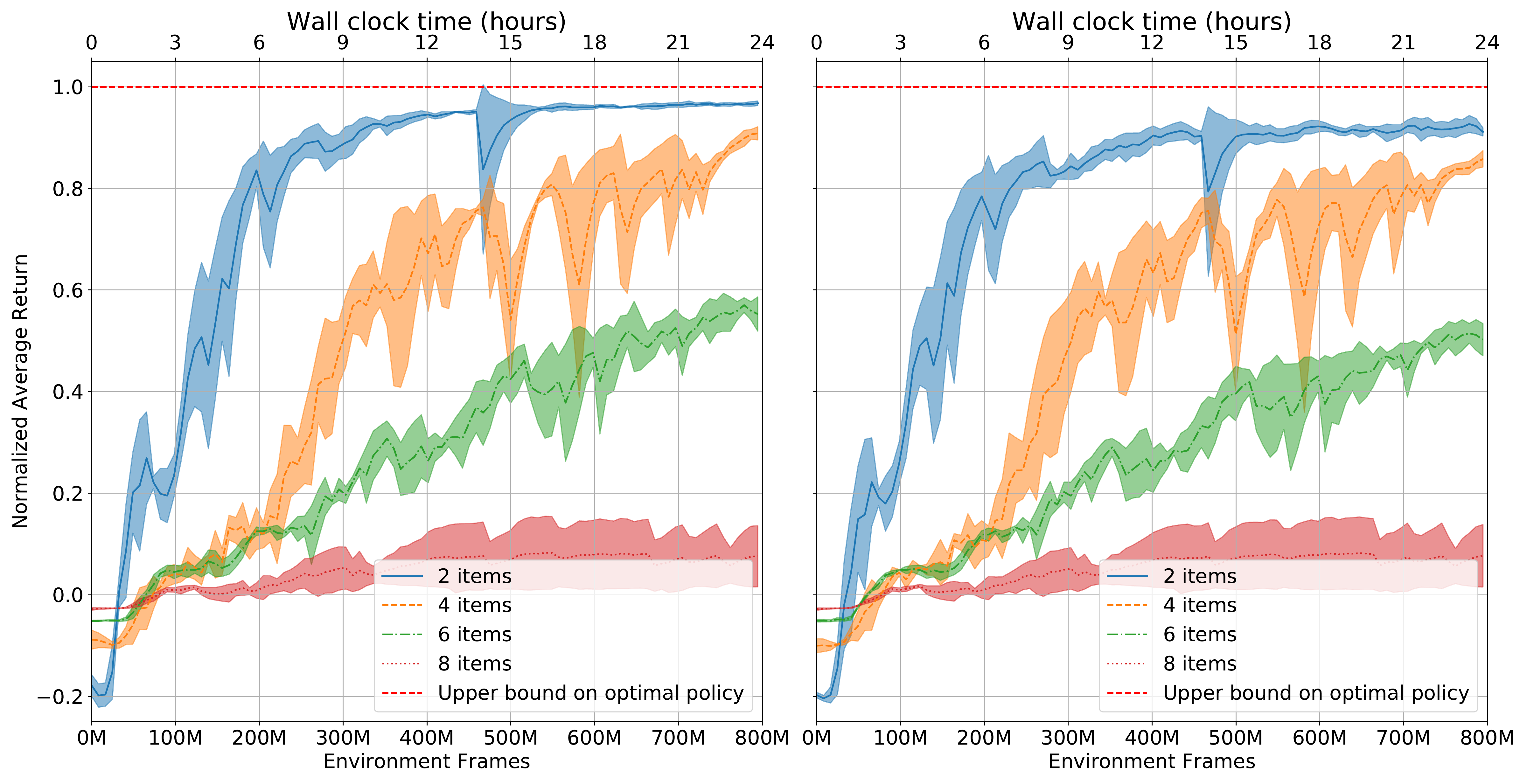}
\end{center}
  \caption{Results from the ''\emph{Ordered k-item}'' scenario for $k\in  [2,4,6,8]$, three experiments were undertaken for each k. Shown are the mean and std.\ of the item-normalized average return of the agent evaluated on the training (left) and testing (right) sets. The 6-item scenario is challenging and the 8-item is too difficult for the baseline agent, these are good scenarios to compare future work.}
\label{fig:kitem_results}
\end{figure}

\noindent
\textbf{Ordered k-item}
Experiments were conducted on 4 different scenarios of a fixed size of 5x5 with 2,4,6 and 8 items, results of the agent's performance are shown in Fig.~\ref{fig:kitem_results}. Here, the more critical parameter is the number $K$ of items to be retrieved in correct order.  For the 2 and 4-item scenario configurations, we observe that the policies plateau near to the optimal policy, the benchmark agent struggles with the 6-item scenario and fails on the 8-item scenario. 

\begin{figure}[t]
\begin{center}
  \includegraphics[width=0.99\linewidth]{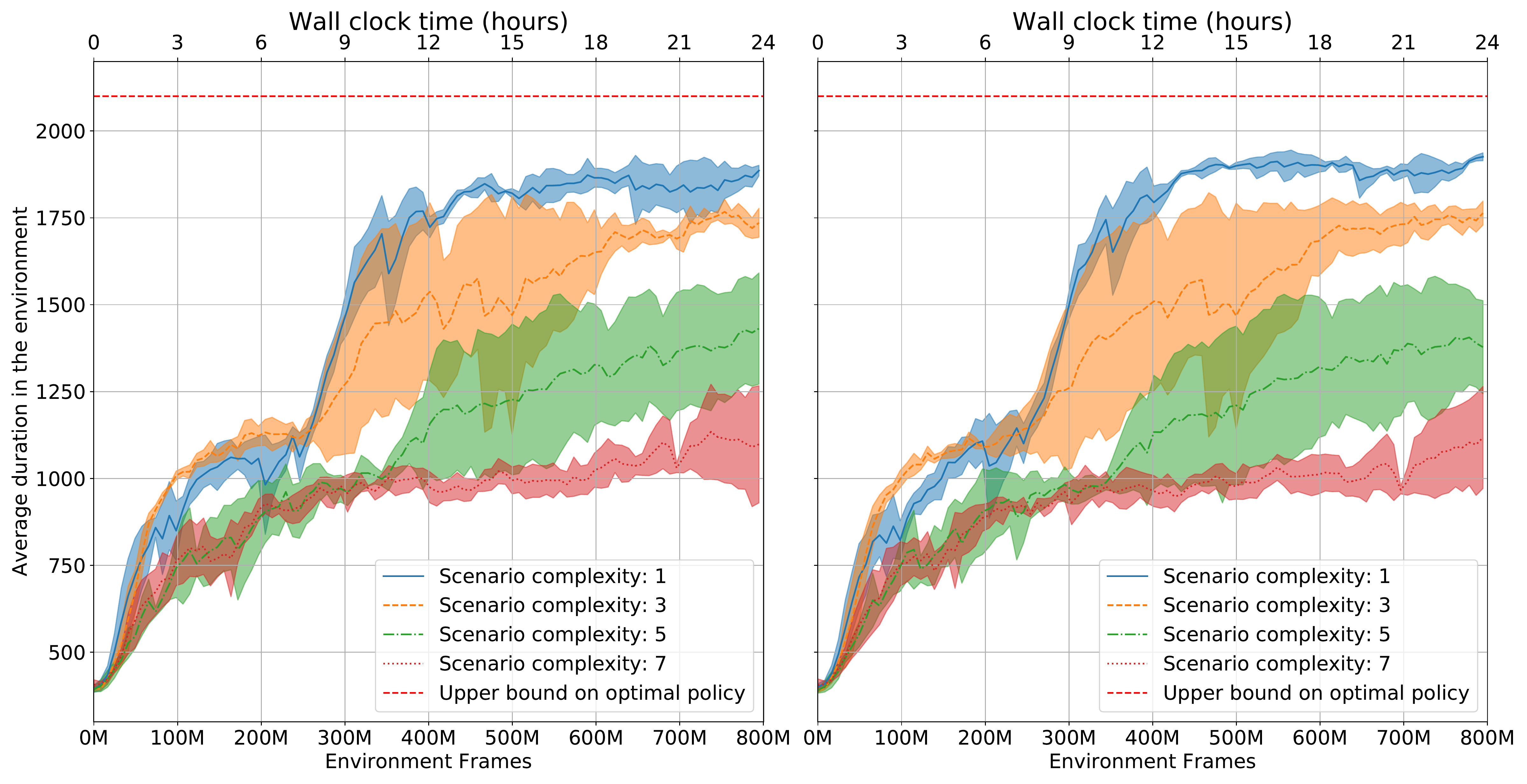}
\end{center}
  \caption{Results from the ''\emph{Two color correlation}'' scenario for four difficulty settings, three experiments were performed for each setting. Shown are the mean and std.\ of the agent's average duration in the environment when evaluated on training (left) and testing (right) datasets.}
\label{fig:twocol_results}
\end{figure}

\noindent
\textbf{Two color correlation:} Experiments were conducted on 4 different scenarios of a fixed size of 5x5 with increasing environment complexity, results of the agent's performance are shown in Fig.~\ref{fig:twocol_results}. We observe for the less complex scenarios with 10 or 30\% of the walls retained, the agent's performance plateaus near the optimal policy. The policies learned on more complex scenarios require further improvement and would be good places to test agent architectures and auxiliary losses.
\begin{figure}[t]
  \begin{minipage}[c]{0.65\textwidth}
    \includegraphics[width=\textwidth]{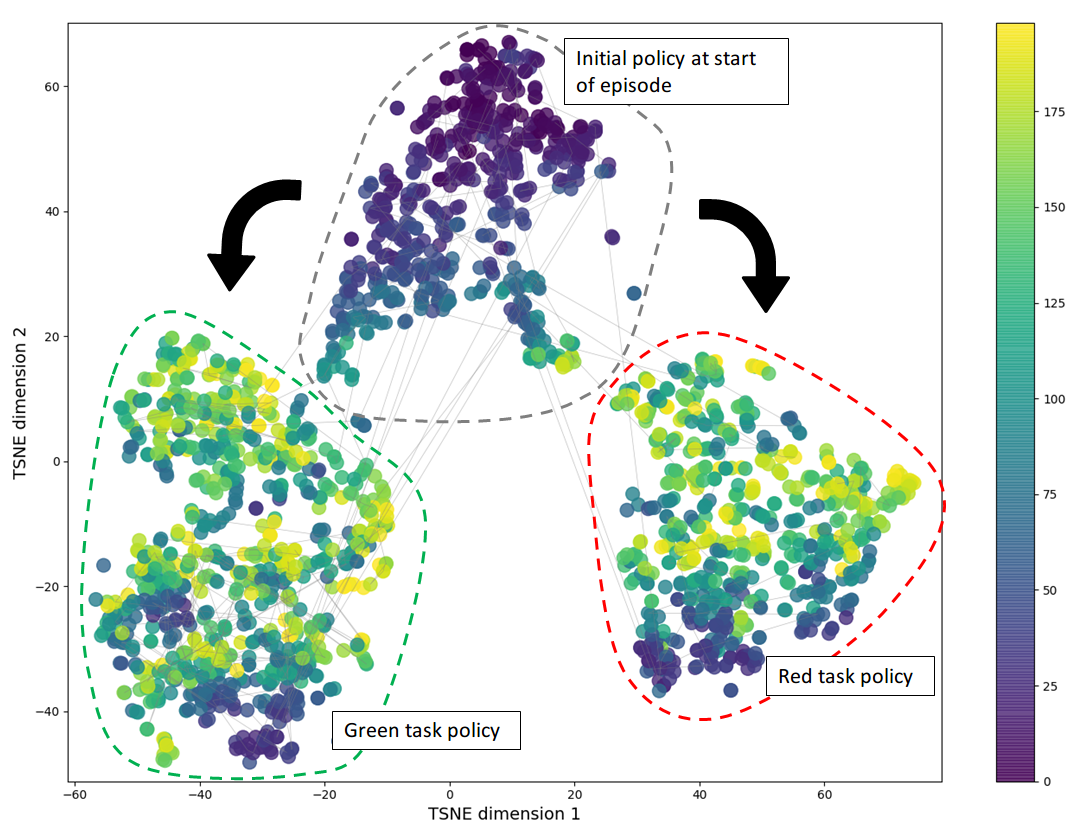}
  \end{minipage}\hfill
  \begin{minipage}[c]{0.34\textwidth}
    \caption{T-SNE analysis of agent's hidden state, sampled from 100 episodes in the ``\emph{Two color correlation}'' scenario. Each point is a 2D representation of one hidden state activation, the color is the time-step, grey lines connect 20 independent trajectories. 
    Once the indicator item has been discovered the policy transitions from an exploration policy to a one of two policies associated with the green task or the red task.
    } \label{fig:tsne}
  \end{minipage}
\end{figure}
\subsection{Analysis and Discussion}
\noindent
To provide insight into the decision making process of the agent, we conducted an analysis of the activations in a trained agent's architecture over episodes in the two color correlation scenario. In Fig.~\ref{fig:abstract_figure}, page \pageref{fig:abstract_figure}, we show an analysis of the progression of the agent's hidden state during the first 64 steps in the environment. 
We observe a large change in the hidden state at step 21, further analysis indicates this is the first time the agent observes the indicator object that identifies the task objective. When the totem is observed there are many boolean flags of the hidden state that are switched on and off, which modifies the agent's behavior policy. This is an example of an agent that has learned to explore, reason and store information, purely from a weak reward signal.

We conducted further analysis by collecting trajectories on hidden states vectors from 100 rollouts of the trained policy, dimensionality reduction was applied using T-SNE \cite{VanDerMaaten2008} on 2000 hidden state vectors selected from a random permutation of the collected trajectories. We identify 3 separate groups of hidden state vectors, shown in Fig. \ref{fig:tsne}. Further analysis indicated that there are three sub-policies that the agent has learned for this scenario; i) an initial policy which explores the environment to identify the color of the task, which transitions to either ii) a policy that collects red items in the case of the red task or iii) a policy that collects green items.

\myParagraph{Challenges in spatial reasoning} the experiments on our proposed suite of benchmarks indicate that current state of the art models and algorithms still struggle to learn complex tasks, involving several different objects in a different places, and whose appearance and relationships to the task itself need to be learned from reward. The difficulty of the proposed scenarios can be adjusted, to scale the task to different levels of required reasoning. For instance, while the ordered $K-$ items scenario is solvable for up to 4 items, it rapidly becomes challenging when the number of items or the size of the environment is increased.
\section{Conclusions}\label{sec:conclusions}
\noindent We have presented four scenarios that require exploration and reasoning in 3D environments. The scenarios can be simulated at a rate that exceeds 12,000 FPS (frame skip=4) on average, including environment resets. Training can be done up to 9,200 FPS (frame skip=4), including forward and backward passes.

Experiments have been conducted for each scenario for a wide variety of difficulty settings. We have shown robust agent performance across random weight initialization for a fixed set of hyper-parameters. Generalization performance has been analyzed on held out test data, which demonstrates the ability of the benchmarks to generalize to unseen environment configurations that are drawn from the same general distribution. We have highlighted limitations of a typical RL baseline agent and have identified suitable scenarios for future research.

Code for the scenarios, benchmarks and training algorithms is available online\textsuperscript{\ref{note1}}, with detailed instructions of how to reproduce this work.



%
%
%
\bibliographystyle{splncs04}
\bibliography{deep_rl_benchmarks.bib}
%




\end{document}